\documentclass{article}
\usepackage{spconf,amsmath,epsfig}
\usepackage{multirow,array}
\usepackage{booktabs}

\title{Instance-aware Remote Sensing Image Captioning\\
with Cross-hierarchy Attention }
\name{Chengze Wang, Zhiyu Jiang$^{*}$, Yuan Yuan\thanks{2020 IEEE. Personal use of this material is permitted. Permission from IEEE must be obtained for all other uses, in any current or future media, including reprinting/republishing this material for advertising or promotional purposes, creating new collective works, for resale or redistribution to servers or lists, or reuse of any copyrighted component of this work in other works. $^*$Corresponding author: Zhiyu Jiang (e-mail: zhiyu.jiang.chn@gmail.com)}}
\address{School of Computer Science and Center for OPTical IMagery Analysis and Learning (OPTIMAL),\\
	Northwestern Polytechnical University, Xi'an 710072, Shaanxi, P. R. China\\}

\begin{document}

%
\maketitle
\begin{abstract}
The spatial attention is a straightforward approach to enhance the performance for remote sensing image captioning.
However, conventional spatial attention approaches consider only the attention distribution on one fixed coarse grid, resulting in the semantics of tiny objects can be easily ignored or disturbed during the visual feature extraction.  
Worse still, the fixed semantic level of conventional spatial attention limits the image understanding in different levels and perspectives, which is critical for tackling the huge diversity in remote sensing images. 
To address these issues, we propose a remote sensing image caption generator with instance-awareness and cross-hierarchy attention.
1) The instances awareness is achieved by introducing a multi-level feature architecture that contains the visual information of multi-level instance-possible regions and their surroundings.
2) Moreover, based on this multi-level feature extraction, a cross-hierarchy attention mechanism is proposed to prompt the decoder to dynamically focus on different semantic hierarchies and instances at each time step.
The experimental results on public datasets demonstrate the superiority of proposed approach over existing methods.

\end{abstract}
\section{Introduction}
\label{sec:Introduction}

Conventional remote sensing image analysis tasks usually focus on the object-level or pixel-level understanding, such as object classification, change detection, and image segmentation. 
Despite that all mentioned tasks have gained massive success and been deployed in many industrials, the lack of comprehensive description to high-level semantics limits the expansion of remote sensing applications.
To access a comprehensive description of global semantic information contained in the captured scene intuitively, image captioning \cite{lu2017exploring, zhang2019multi,zhang2019description} is introduced to the remote sensing field, which generates sentences in human language to summarize the high-level semantic content from remote sensing images. 

Image captioning is an interdisciplinary task emerged from the overlap of computer vision and natural language processing, and has aroused great attention from the community. 
It is complicated to accomplish automatically, since it not only requires a global understanding of objects and their relations, but also needs to transform the visual content into flexible and fluent sentences. 
Early image caption approaches were mainly template-based models \cite{li2011composing} and retrial-based models \cite{ordonez2011im2text}, have been replaced by encoder-decoder based methods  \cite{vinyals2015show} for better performance and flexibility. 
Later on, to further enhance the correlation between the visual information and generated words during the encoding-decoding progress, the visual attention mechanism \cite{xu2015show} was explored. 
Nowadays, most of image captioning approaches are designed based on the attention-enhanced encoder-decoder architecture. 

When comes to remote sensing image captioning, there are several published works that have explored this field in recent years. 
The first attempt was made by Qu \emph{et al.} \cite{qu2016deep} who utilized an encoder-decoder framework based on multimodal neural networks to prove the possibility of generating human sentences for remote sensing images. 
The attention mechanism for remote sensing image was firstly introduced in \cite{lu2017exploring}, and found out that the spatial attention is of crucial importance for captioning performance. 
Moreover, a multi-scale cropping mechanism \cite{zhang2019multi} was designed to adapt different size instances in the image.
Lately, Zhang \emph{et al.} \cite{zhang2019description} introduced the classification label of remote sensing images into attention mechanism, which improved the performance considerably.


\begin{figure*}[tbp] 
\centering 
\label{fig1}
\includegraphics[width=18cm]{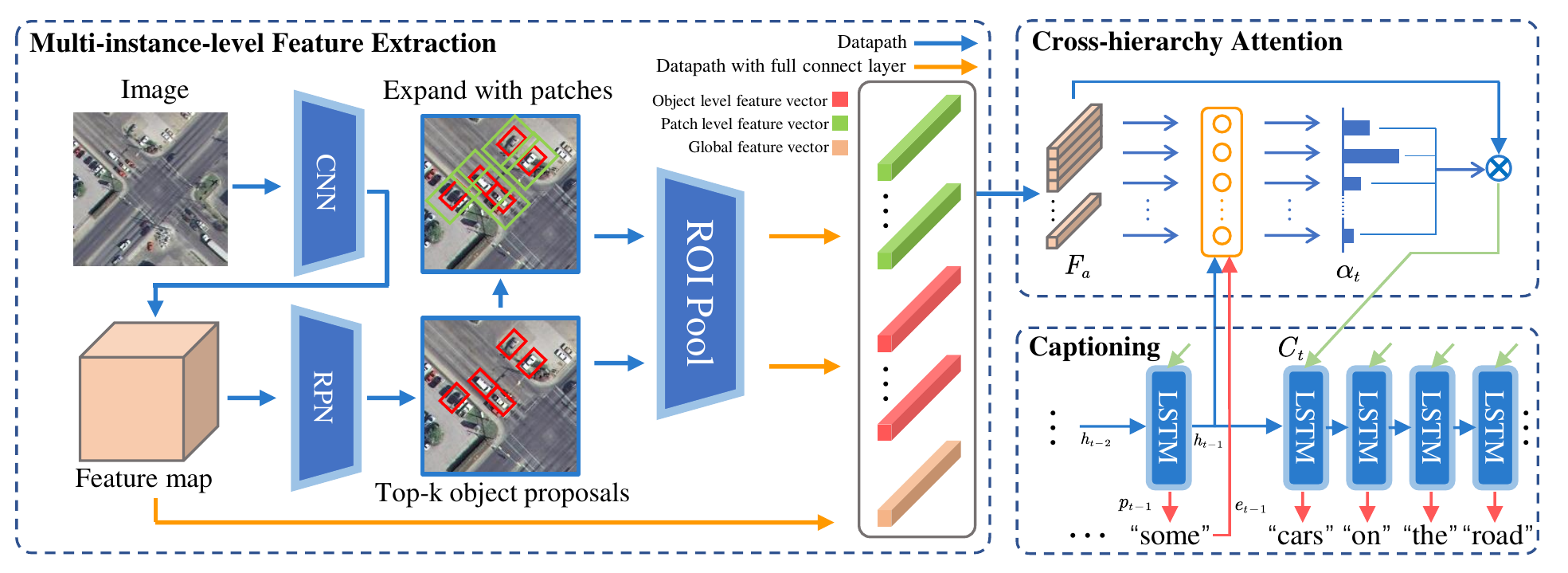} 
\caption{The overall architecture of the proposed method.
The model takes remote sensing images to the multi-instance-level feature extraction as encoding, then outputs descriptive sentences with the cross-hierarchy attention mechanism.
} 
\vspace{-0.6cm}
\end{figure*}

Though previous remote sensing image captioning methods have gained remarkable progress, there are still existing limitations as follows: 
1) Mostly, the spatial attention is computed on the feature map with a low spatial-wise resolution grid, which leads to that the encoder is difficult to distinguish content with accurate, clear boundaries.
At the same time, this grid-based attention is difficult to focus on tiny or irregular ground targets, causing the attended semantic feature usually containing quite a lot irrelevance visual information.
2) The attending standpoint is fixed to a particular scale and semantic level. 
However, in remote sensing image captioning, part of captured scenes are with extremely brief (\emph{e.g.} ocean, desert) or extremely complex semantic information (\emph{e.g.} city park, highway), and the fixed scale and semantic level will make it difficult to obtain a reasonable attention distribution.

Considering the limitations mentioned above, an instance-aware remote sensing image captioning method with cross-hierarchy attention is proposed.
The contributions of this paper are summarized as follows:

1) Instance-awareness is for the first time realized in remote sensing image captioning. 
It is a more straightforward way to distinguish semantic information with the ground objects and their relations.
The extracted object feature together with the neighboring patch feature and global feature constitutes the multi-instance-level feature, which improves the accuracy and possibility of feature attention at the spatial and semantic hierarchies.

2) We propose a cross-hierarchy attention mechanism to accommodate the multi-level feature inputs, prompting the decoder to dynamically focus on different semantic levels and different instances. 
This enhances the flexibility on semantics and scale when facing extreme remote sensing scenes.

The evaluation on mainstream datasets demonstrates the effectiveness of the proposed method.

\section{Methodology}
\label{sec:Methodology}
\subsection{Multi-instance-level Feature Extraction}
Following the basic encoder-decoder architecture, the framework of our proposed approach is shown in Fig.\ref{fig1}. 
As the encoder part of the whole image caption generator, a multi-instance-level feature extractor is designed. These encoders simultaneously aggregate 3 different levels of instance visual feature, including the object-level feature, the patch-level feature and the global feature. 
After extraction, these features will be fed into the cross-hierarchy attention mechanism, achieving a more accurate and reasonable spatial focus on the feature when predicting words.

\subsubsection{Object-level Feature Extraction}
For describing most scenarios, salient ground objects are the key to description generation.
To accurately locate salient targets in a remote sensing image, we select the top-$n$ ROIs nominated by the Faster RCNN \cite{ren2015faster} as $n$ possible regions of key objects. 
To minimize the gap between different datasets, the class-agnostic pre-trained model is adopted.
ROI-pooling and full connect layers are applied to each proposed ROI so that the output feature vector sequence $F_o$ has the same dimension for each object:
$
F_o=[ F^1_o ,F^2_{o},\dots,F^n_{o} ] .
$

\subsubsection{Patch-level Feature Extraction}
Remote sensing image caption is highly relative to the main objects and their spatial relations, which means the semantic information of the neighboring part of one salient object is as important as the object itself.
On the other hand, a larger perception field enhances the robustness of visual feature extraction, especially for the inaccuracy caused by the region proposal module.
In our approach, the patch of each object is defined as the rectangular area that has the same ratio, center, and direction of the responding object, but the scale of the patch is determined by a scaling-factor $k$. If the patch bound exceeds the image edge, the patch region is cropped to fit the image size. With the same after-ward processing as the object-level ROIs, the feature of multiple patches $F_p$ is extracted. 

\subsubsection{Global Feature Extraction}

In scenarios lack of texture diversity (\emph{e.g.} ocean, forest, and desert), so an independent global feature extraction branch is involved to provide a more comprehensive feature hierarchy. 

Unlike other spatial-attention-based methods, our above mentioned two feature extraction levels have brought the visual information of key objects and their surroundings to the decoder, so the global feature is the feature vector output by full connection layer instead of the feature cube acquired from the end of convolutional layers, which also boosts computation efficiency of the decoder at the same time.
Specifically, we utilize the output of the last full-connect layer of a Resnet-101 as the global feature, which has the same dimension of $F_p$ and $F_p$, and is denoted as $F_g$.
At last, the overall feature $F_a$ is the stack of all three level features:
\begin{equation}
\begin{array}{l}
F_a=[  F^1_o,\dots ,F^n_{o},F^1_{p},\dots,F^n_{p},F_g ].
\end{array}
\end{equation}

\subsection{Cross-hierarchy Attention and Caption Generation }

The proposed cross-hierarchy attention mechanism is similar to the spatial attention mechanism, while having a higher diversity of focusable scale and semantic levels. The core idea of spatial attention is using the former state of LSTM to decide the next focus point or area on a uniform feature map grid. With the more oriented image feature as the input of each word's prediction, the performance and interpretability of image caption have been significantly improved.

Following this basic way, the cross-hierarchy attention dynamically re-weight the input multi-instance-level feature to focus on different area or the whole image at each time step. 
In the cross-hierarchy attention mechanism, the extracted feature vector of each instance can be regard as the feature of a cell in conventional spatial attention grid. 


More specifically, given a stacked multi-level feature $F_a$, each of which is $d$ dimensional, $F_a$ is fed as a concatenation with the text feature and the hidden state on the last time step to a single layer neural network to calculate the attention score map over different levels and instances,
\begin{equation}
\begin{array}{l}
{a}_{t}=w_{h}^{T}\tanh(W_{a}[{F_{a},h_{t-1},e_{t-1}}] ),
\end{array}
\end{equation}
where $W_a$,  $w_h$ are network weights to be trained. 
$e_{t-1}$ is the text feature vector, which is extracted from the last generated word by embedding its one-hot feature into same $d$ dimensional.
And the hidden state $h_{t-1}$ comes from the LSTM outputs at last time step,
\begin{equation}
\begin{array}{l}
h_{t-1}=LSTM(C_{t-1}, h_{t-2}),
\end{array}
\end{equation}
where $C_{t-1}$ is the corresponding derived image feature. 
 
Then, the attention distribution mask $\alpha_{t}$ is computed by a softmax layer,
\begin{equation}
\begin{array}{l}
\alpha_{t}^{i}=\exp \left(a_{t}^{i}\right) / \sum_{j}^{n} \exp \left(a_{t}^{j}\right).
\end{array}
\end{equation}
At the beginning of each generation, the attention distribution is uniformly initialized. 
Subsequently, based on this attention distribution $\alpha_{t}^{i}$, the cross-hierarchy attention derived image feature $C_{t}$ is computed as follows,
\begin{equation}
\begin{array}{l}
C_{t}=\sum_{i=1}^{N} \alpha_{t}^{i} F_{i}.
\end{array}
\end{equation}
At time step $t$, given the attended feature $F_{d}$, hidden state $h_{t-1}$, the probability of word prediction $p_{t}$ is
\begin{equation}
\begin{array}{l}
p_{t}=\operatorname{softmax}\left(U_{p} \tanh \left(W_{p}\left[C_{t}, h_{t-1}\right]+b_{p}\right)\right).
\end{array}
\end{equation}

At last, the $t$-th word of the description is chosen from the pre-embedding vocabulary vector by the maximum probability from $p_{t}$.

\section{Experiments}
\label{sec:Experiments}
\begin{table*}[ht]
\centering
\scriptsize
\caption{Evaluation results on UCM, Sydney and RSICD dataset. Bold represents the top one.}
\label{Tab03}
    \begin{tabular}{p{1.6cm}|p{0.4cm}p{0.4cm}p{0.4cm}p{0.4cm}p{0.4cm}p{0.5cm}|p{0.4cm}p{0.4cm}p{0.4cm}p{0.4cm}p{0.4cm}p{0.5cm}|p{0.4cm}p{0.4cm}p{0.4cm}p{0.4cm}p{0.4cm}p{0.5cm}}
    \toprule
      Dataset    & \multicolumn{6}{c}{UCM}                       & \multicolumn{6}{c}{Sydney}                    & \multicolumn{6}{c}{RSICD} \\
          \midrule
       Criteria   & B-1 & B-2 & B-3 & B-4 & C & R & B-1 & B-2 & B-3 & B-4 & C & R& B-1 & B-2 & B-3 & B-4 & C & R\\
          \midrule
    Zhang \emph{et al.} \cite{zhang2019multi} & 0.594 & 0.532 & 0.481 & 0.429 & /     & /     & {0.615} & {0.540} & {0.473} & {0.400} & {/} & {/} & {/} & {/} & {/} & {/} & {/} & {/} \\
    Attention \cite{lu2017exploring} & 0.745 & 0.655 & 0.586 & 0.525 & 2.612 & 0.724 & {0.732} & {0.667} & {0.622} & {0.582} & \bf{2.499} & {0.713} & {0.676} & {0.531} & {0.433} & {0.36} & {1.964} & {0.611} \\
    FC-ATT \cite{zhang2019description}& 0.814 & 0.750  & 0.685  & 0.635 & 2.996 & 0.750 & {0.808} & {0.716} & {0.628} & {0.554} & {2.203} & {0.711} & {0.746 } & {0.625 } & {0.534 } & {0.457} & \bf{2.366} & {0.633} \\
    SM-ATT \cite{zhang2019description}& 0.815 & 0.758 & 0.694 & 0.646 & 3.186 & \bf{0.763} & {0.814} & {0.735} & \bf{0.659} & {0.580} & {2.302} & {0.719} & {0.757} & { 0.634 } & \bf{0.538 } & {0.461 } & {2.356} & {0.646} \\
    \midrule
    Proposed & \bf{0.823} & \bf{0.768}  & \bf{0.710}  & \bf{0.659}  & \bf{3.192}  & 0.756 &   \bf{0.817}   &  \bf{0.742}  &  0.657  &  \bf{0.591}    &  2.291  &\bf{ 0.721}   & \bf{ 0.770}  &\bf{ 0.649}   &  0.532 &  \bf{0.471}   &  2.363  & \bf{0.651} \\
\bottomrule
\end{tabular}
\end{table*}
 
\subsection{Datasets and Evaluation Metrics}

We adopt all
 3 remote sensing image caption datasets to our quantitative evaluation. Each dataset is divided into three parts by a default 80\%:10\%:10\% ratio, which is for training, evaluation and test, respectively.

{\bf UCM-Captions} \cite{qu2016deep} is a mid-scale remote sensing image caption dataset. This dataset is originally used for scene classification, and then extended with manual descriptions to become the most popular caption dataset. There are 21 categories of scenes and 100 images for each category, and each image contains 5 sentences as description.

{\bf Sydney-caption} \cite{qu2016deep} is a relatively small-scale remote sensing image caption dataset, like UCM-Caption, it is also extended from a scene classification dataset. The dataset includes 613 pictures, each with 5 sentences as captions.

{\bf RSICD} \cite{lu2017exploring} is nowadays the most complicated remote sensing image caption dataset because of the largest scale and high difficulty. RSICD includes 10,921 pictures, but there are only 24,333 ground truth descriptions. For incompletely described images, copies of existing sentences are used to meet the model input requirements of 5 sentences per picture.

Since the manual estimation of quality and accuracy of the generated sentences is susceptible to subjective influences, a series of quantitative evaluation criteria from natural language processing are introduced to our quantitative evaluation. They are BLEU-n (n=1,2,3,4), CIDEr, and ROUGE-L, which abbreviated as B-n, C and R in Table \ref{Tab03}. 

\subsection{Implementation Details and Results}

During the training, the self-critical training mechanism \cite{rennie2017self} and beam search are both utilized with the beam size of 2.
For the salient object region proposal, a Faster RCNN with Resnet-101 backbone is pre-trained on MSCOCO detection datasets while class-agnostic is enabled.
For global feature extraction, another Resnet-101 model pre-trained on ImageNet classification dataset is chosen to conduct comprehensive semantic information. 
The caption decoder is trained under the cross-entropy objective using ADAM optimizer with the learning rate of a fixed $10^{-4}$.
The patch scale factor $k$ is fixed to 2.0 during the test, and the number of detection objects $n$ is set to 5.

Table \ref{Tab03} shows the quantitative results of various approaches on 3 datasets. It should be noted that the listed approaches utilized different CNN encoder architecture to extract features. 
To be specific, ``Attention" \cite{lu2017exploring}, ``FC-ATT" \cite{zhang2019description} and ``SM-ATT" \cite{zhang2019description} utilized VGG16 as backbone model, Zhang \emph{et al.} \cite{zhang2019multi} utilized ResNet-152, and all of them were pre-trained on ImageNet. 
However, the utilized object detector in our implementation was pre-trained on MSCOCO, which is not a remote sensing related dataset, the gap of different datasets definitely draws our performance back. 

It can be observed from Table 1, the comparison results indicate that our proposed method surpasses the mainstream methods at most metrics in all datasets. 
Practically, on the BLEU-n series metrics, our method has shown its superiority most. while the proposed method is much more likely to lag behind on CIDEr and ROGUE-L by other attention-based methods. 
\begin{figure}[t] 
\centering 
\includegraphics[width=8.6cm]{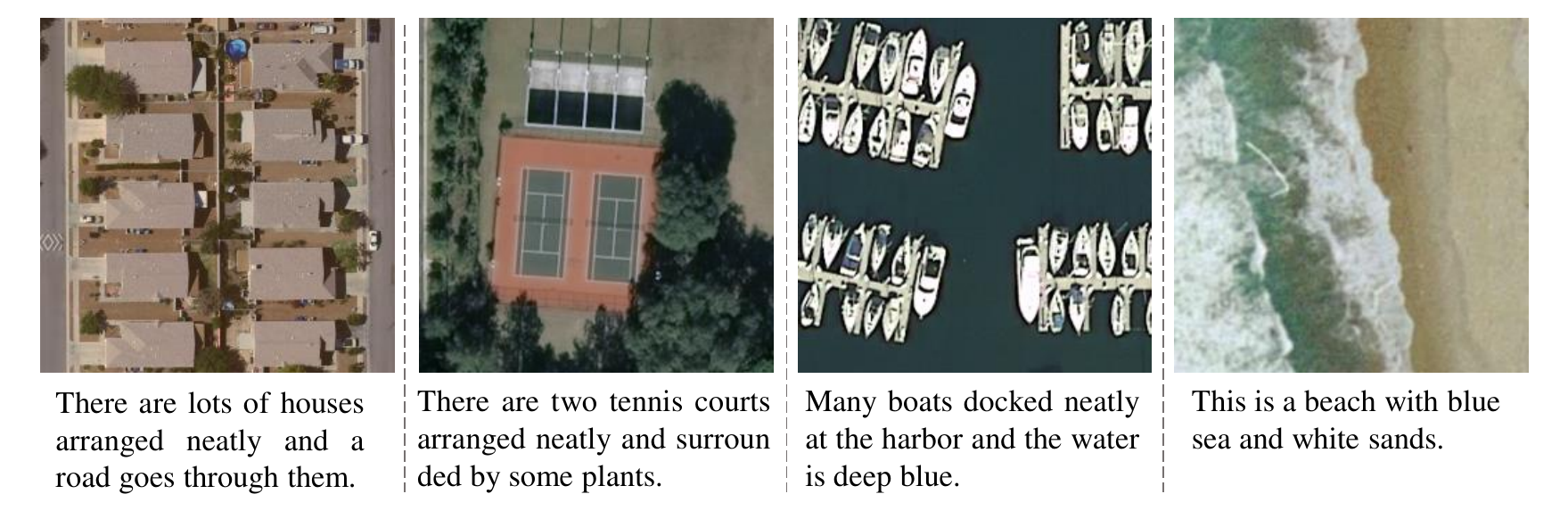} 
\vspace{-0.6cm}
\caption{Example results of our proposed method.
\vspace{-0.6cm}
} 
\end{figure}
\section{Conclusion}
\label{sec:Conclusion}
In this paper, we propose a remote sensing image caption method with instances-awareness and cross-hierarchy attention. 
In this work, the Faster RCNN is utilized to accurately locate the key objects and their surroundings.
In this way, the encoder is capable of extracting key element regions accurately, instead of estimating the content based on the uniform spatial grid. 
To deal with single-textured scenes like deserts and oceans, a global visual feature is also introduced.
To adapt to this multi-instance-level feature form, a cross-hierarchy attention mechanism is proposed to prompt the decoder dynamically focus on different semantic levels and different instances of visual feature. 
The experimental results on mainstream datasets show the effectiveness of the proposed approach.

\bibliographystyle{IEEEbib}
\small{\bibliography{refs}}

\end{document}